\documentclass[11pt]{article}

\usepackage[final]{acl}

\usepackage{times}
\usepackage{latexsym}
\usepackage{booktabs}
\usepackage{paralist}
\usepackage{array}
\usepackage{ragged2e}
\usepackage{hyperref}
\newcolumntype{L}[1]{>{\RaggedRight\arraybackslash}p{#1}}
\newcolumntype{R}[1]{>{\RaggedLeft\arraybackslash}p{#1}}
\renewcommand{\arraystretch}{1.08}

\usepackage[T1]{fontenc}
\usepackage[utf8]{inputenc}

\usepackage{microtype}

\usepackage{inconsolata}

\usepackage{graphicx}

\title{Two Centuries of Sexism in British Parliament: A Computational Analysis of Women's Representation in the Hansard Corpus}

\author{
  Mohammad Omar Khursheed\textsuperscript{1}\thanks{Mohammad Omar Khursheed and Mandira Sawkar are equal-contribution first authors. Ashiqur R. KhudaBukhsh is the corresponding author.} \and
  Mandira Sawkar\textsuperscript{2}\footnotemark[1] \and
  Ashiqur R. KhudaBukhsh\textsuperscript{2}
  \\
  \textsuperscript{1}Independent \quad
  \textsuperscript{2}Rochester Institute of Technology \\
  \texttt{omarkhu@gmail.com} \quad
  \texttt{ms7201@rit.edu} \quad
  \texttt{axkvse@rit.edu}
}
\begin{document}
\maketitle
\begin{abstract}
The language a legislature uses to debate women's rights, even in favour of them, encodes systematic patterns of sexism that persist across two centuries. In this work, we analyse 6,531 speeches over 200 years of UK parliamentary debate (Hansard, 1803-2005) by using large language models to classify a speaker's perspective towards women's suffrage and political representation, as well as analyse sexist speech in parliament from the lens of the Ambivalent Sexism Inventory. We also release this parliamentary dataset, an organized and metadata-enriched version of the publicly available Hansard Corpus optimized for computational social science research, with 6.7 million speeches across 1.2 million debates, with 89\% gender-matching for speeches by MPs from the House of Commons. We find that 54\% of speeches opposing women's representation contain sexist content, compared to 21\% of speeches that are for the cause, and that the two sides use fundamentally different types of sexism: anti-suffrage rhetoric combines hostile and benevolent framing, while pro-suffrage sexism is overwhelmingly benevolent. Female MPs support women's political rights at 93\% compared to 70\% for male MPs, a gap that closes only after enfranchisement. Our findings are evidence that benevolent and hostile sexism are used in different rhetorical contexts in a manner consistent with the theory of Ambivalent Sexism.
\end{abstract}

\section{Introduction}

Parliamentary debate in Britain has, since the creation of the institution in 1801, served a role as a driver of policy that directly affects entire nations through decisions made by a relatively small number of elected (and unelected) representatives. The rich historical web record of debate transcripts, therefore, serves as a reflection of societal evolution and its availability through the Hansard archive serves as the basis of this work.

Women in 19th century Britain had no political representation. This began to change in 1918 with two Acts of Parliament: the Representation of the People Act, which gave women over 30 (with property qualifications) the right to vote, and the Parliament (Qualification of Women) Act, which allowed women to stand for Parliament \citep{representation1918act_parliament, representation1918act_journal, Taylor_2022}. In 1928, the Representation of the People (Equal Franchise) Act extended voting rights to women on equal terms with men (age 21+) \citep{franchise1928act_parliament, Taylor_2022}. These legislative changes were preceded and accompanied by extensive parliamentary debate, which forms the focus of our analysis.

The suffrage movement, the campaign for women's right to vote and stand for political office, is well-documented by historians~\citep{van2016women,holton2002suffrage}, but computational analysis of the actual arguments used is limited. In this paper, we investigate whether proponents and opponents of suffrage employed fundamentally different types of reasoning, or whether they used similar rhetorical strategies to reach opposing conclusions. Understanding these discourse patterns reveals how political disagreement operates: whether opposing sides differ in the premises they start from or in the type of gendered reasoning they deploy.

We rely on the framework of the Ambivalent Sexism Inventory \citep{glick1996} to ground our computational analysis in the relevant sociological literature. This taxonomy explains how prejudice against women can be viewed as a combination of hostile sexism, referring to overtly negative and antagonistic attitudes which characterize women as inferior, and benevolent sexism, where often chivalrous and flattering arguments are applied to cast women as frail and helpless.

We analyse 6,531 suffrage-related speeches drawn from 6.7 million speeches in the Hansard Parliamentary Corpus (1803-2005) and classify the speaker's stance towards women's political representation using LLMs. We further classify whether the speech contains sexist elements and whether they are hostile or benevolent. We validate the quality of LLM-based classification on both axes of stance and sexism against human annotations on a 300-speech validation set. 

This paper makes the following contributions. First, we provide a large-scale, gender-matched dataset of 6.7 million British parliament speeches~\footnote{Our dataset and reproducible code are available at \url{https://github.com/Social-Insights-Lab/Two-Centuries-of-Sexism}} enabling future computational social science research. Second, we classify sexism in historical political discourse under the Ambivalent Sexism Inventory \citep{glick1996}, providing naturalistic evidence for a framework developed in laboratory settings and demonstrating that careful application of LLMs as judges can yield meaningful insights on temporally varied social science data. Third, we show that opposing political stances deploy fundamentally different types of gendered reasoning, not merely different conclusions, connecting computational analysis to theoretical frameworks grounded in the social science literature.

\section{Related Work}

Prior work on parliamentary corpora has focused on sentiment, stance, and ideological scaling, while gender-focused studies examine style and representation. These rarely analyse the type of sexism deployed. Our work bridges this gap by classifying sexism in Hansard along established social-psychological axes.

\noindent\textbf{Parliamentary Debates Analyses.} NLP work on parliamentary corpora has focused on sentiment, stance, and ideological scaling. Prior work \citep{Abercrombie2020-lit-rev, Abercrombie2018, abercrombie-batista-navarro-2020-parlvote, abercrombie-batista-navarro-2022-policy} reviews sentiment and stance analysis in legislative text and introduce sentiment-labelled Hansard resources such as ParlVote, enabling transformer-based sentiment and stance classification over UK debates, while traditional scaling models such as Wordfish infer ideological positions from word frequencies in parliamentary corpora \citep{slapin-etal}. More recent work extends these ideas to other legislatures and tasks, including sentiment and topic classification in German parliamentary speeches \citep{Alexander-etal}, deep-learning sentiment models tailored to parliamentary data in ParlaSent \citep{Mochtak31122025}, and argument-aware summarisation evaluated on UK parliamentary datasets \citep{Altameemi-etal}. Beyond UK parliamentary discourse, notable works exist on longitudinal US political speeches such as the recent analyses of immigration issues by ~\citet{card2022computational}. 

\noindent\textbf{Gender, Suffrage, and Political Speech.} Gender-focused work on parliamentary discourse has examined style, topics, and representation. This includes work that shows that women’s historically more ``feminine'' styles (emotional, conciliatory) have converged toward traditionally ``masculine'' argumentative styles over time \citep{Hargrave_Blumenau_2022}; attempt to use text mining on Hansard to study whether women have changed the substantive content of debate and the representation of ``women’s issues'' \citep{Blaxill-etal}; and others find gendered patterns in topics and styles without pre-defining women’s vs men’s issues \citep{Anagnosticapproachtogender}. Beyond Westminster, computational feminist standpoint analyses examine how ``gender'' is framed in EU institutions \cite{Yılmaz06102025}, and diachronic studies of British parliamentary discourse trace shifts in gendered and gender-neutral forms \citep{Soriano-Jiménez_2024}.

\noindent\textbf{Stance Detection in Political Text.} Stance detection in parliamentary debates has been used to detect ethotic appeals and their polarity across UK Commons debates \citep{Duthie-etal} and combined textual and acoustic features for claim detection in televised political debates \citep{Lippi_Torroni_2016}. Work on online debates shows that explicit argument structure improves persuasion prediction \citep{li-etal-2020-exploring-role}.

\noindent\textbf{Sexism in NLP.} Gender bias studies using NLP have been widely documented and studied in rich and diverse settings~\cite{kumar2020nurse,khadilkar2022gender}. Many datasets annotating sexism in online content have been published \citep{guest-etal-2021-expert, zeinert-etal-2021-annotating}. Classifying types of sexism on online content has been studied \citep{kirk-etal-2023-semeval, jha-mamidi-2017-compliment}. Further lines of work focus on gender bias specifically in LLMs include analysing gender stereotypes in large models \citep{Kotek_etal}; evaluating gender bias in ChatGPT-style systems across multiple languages \citep{ding-etal-2025-gender}; and systematically testing GPT-4–class models on occupational-role scenarios to quantify gendered responses \citep{info16050358}.

\section{Hansard Data}

We collected 1,197,828 debates from the Hansard Parliamentary Corpus \citep{HistoricHansard}, from which we extracted 6,783,015 individual speeches by 52,661 speakers across 203 years (1803-2005) of House of Lords and House of Commons proceedings. We restrict our analysis to the House of Commons, where we achieved 89.3\% gender-matching coverage compared to only 1.2\% in the House of Lords. The Commons subset contains 5.6 million speeches from 12,395 male MPs (4.8M speeches) and 254 female MPs (150,867 speeches).

We matched speakers to individual MPs using a cascaded approach combining ministerial titles, constituency records, temporal matching, and fuzzy string matching with Levenshtein distance thresholds. We prioritised precision over recall to prevent misgendering. This process achieved 89.3\% coverage (see Appendix~\ref{sec:gender_matching} for details).

We identified suffrage-related speeches using a two-tier keyword search applied to speech text, following prior work that uses keyword-based filtering as a high-recall approach to corpus construction \citep{halterman2021corpus, dutta2022murder}. Tier 1 matches explicit suffrage terms (e.g., ``women's suffrage'', ``votes for women''); Tier 2 matches \textit{women/female} in proximity to voting-related terms. The full keyword list and precision estimates are in Appendix~\ref{sec:keywords}. This yielded 6,531 speeches spanning 1809-2004, with 2,307 unique speakers (86.6\% male, 6.6\% female, 8.0\% unknown gender). The concentration of suffrage debate occurred during 1900-1935, bracketing major legislative milestones of 1918 (The Representation of the People Act) and 1928 (The Equal Franchise Act).

Women were first permitted to stand for election to Parliament under the Parliament (Qualification of Women) Act 1918. Nancy Astor became the first woman to take her seat in the House of Commons in November 1919, delivering her first recorded speech on 1 March 1920. Table~\ref{tab:dataset_stats} presents key statistics of our dataset.

\begin{table}[t]
\centering
\footnotesize
\begin{tabular}{@{}L{0.48\columnwidth}R{0.40\columnwidth}@{}}
\toprule
Metric & Value \\
\midrule
Total debates & 1,197,828 \\
Total speeches & 6,783,015 \\
Total speakers & 52,661 \\
Date range & 1803-2005 \\
\addlinespace[3pt]
\multicolumn{2}{@{}l}{\textbf{House of Commons}} \\
\midrule
Total speeches & 5,575,783 \\
Gender-matched speeches & 4,980,711 (89.3\%) \\
Male MPs (with speeches) & 12,395 \\
Female MPs (with speeches) & 254 \\
Male speeches & 4,829,844 \\
Female speeches & 150,867 \\
\addlinespace[3pt]
\multicolumn{2}{@{}l}{\textbf{Suffrage-related speeches}} \\
\midrule
Total suffrage speeches & 6,531 \\
Date range & 1809-2004 \\
Unique speakers & 2,307 \\
Male speakers & 1,998 (86.6\%) \\
Female speakers & 152 (6.6\%) \\
Unknown gender speakers & 185 (8.0\%) \\
Male speeches & 5,430 (83.1\%) \\
Female speeches & 611 (9.4\%) \\
Unknown gender speeches & 490 (7.5\%) \\
\bottomrule
\end{tabular}
\caption{Hansard dataset statistics.}
\label{tab:dataset_stats}
\end{table}

\section{Taxonomy}

We introduce a dual-axis annotation taxonomy for characterizing suffrage-related parliamentary speeches along two dimensions: stance toward women's suffrage and the presence and type of sexism. 

\subsection{Stance}

Each speech is annotated for its expressed position toward women's suffrage and, more broadly, the participation of women in public and political life. We define four mutually exclusive categories:

\begin{compactitem}
    \item \textbf{For:} The speaker explicitly or implicitly advocates increasing women's representation in political institutions and public life.
    \item \textbf{Against:} The speaker explicitly or implicitly opposes women's broader participation in the public sphere.
    \item \textbf{Both:} The speech contains mixed or ambivalent sentiments, expressing partial support alongside reservations or contradictory positions on the topic.
    \item \textbf{Irrelevant:} The speech does not substantively engage with the topic of women's political representation or public life. This category serves as a filtering mechanism to identify false positives introduced by the keyword-based retrieval pipeline.
\end{compactitem}

\subsection{Sexism}

We adopt \citet{glick1996}'s Ambivalent Sexism Inventory (ASI) as the operational framework for annotating sexism in Hansard debates, bridging qualitative social psychology and computational analysis. The ASI posits that sexism comprises two distinct but connected forms: Hostile Sexism (HS) and Benevolent Sexism (BS). HS reflects overtly negative attitudes toward women, particularly those perceived as challenging male authority or traditional gender roles. BS, while superficially positive, constrains women by casting them as pure, fragile, and requiring male protection. This distinction is critical for political discourse, where exclusion can be justified through either direct hostility or paternalistic deference, strategies a binary sexist/non-sexist label would conflate.

 \citet{glick1996} further decompose HS and BS along three domains (paternalism; gender differentiation; and heterosexuality) each with a hostile and benevolent variant. Under HS, these manifest as dominative paternalism (``women should be controlled''), competitive gender differentiation (``women are incompetent, irrational, etc.''), and heterosexual hostility (``women are manipulative or power-seeking''). Under BS, the corresponding forms are protective paternalism (``women should be cherished and protected by men''), complementary gender differentiation (``women possess complementary qualities like purity, morality, etc.''), and heterosexual intimacy (``man is not complete without the love of a woman''). Though these forms differ in tone and logic, they reinforce the same hierarchical structure.
Table~\ref{tab:taxonomy} summarizes the categories used.

\begin{table}[t]
\centering
\footnotesize
\renewcommand{\arraystretch}{1.2}
\begin{tabular}{@{}L{0.45\columnwidth}L{0.45\columnwidth}@{}}
\toprule
\textbf{Hostile} \textit{(degrading, controlling)} & \textbf{Benevolent} \textit{(idealising, restricting)} \\
\midrule
\textit{\textbf{Dominative paternalism}}: women require male authority
& \textit{\textbf{Protective paternalism}}: men should care for and shield women \\
\textit{\textbf{Competitive gender differentiation}}: men are more competent in public life
& \textit{\textbf{Complementary gender differentiation}}: women have distinct purity, moral sensitivity, and nurturance \\
\textit{\textbf{Heterosexual hostility}}: women use sexuality manipulatively
& \textit{\textbf{Heterosexual intimacy}}: men are incomplete without a woman as a romantic partner \\
\bottomrule
\end{tabular}
\caption{Ambivalent sexism taxonomy \citep{glick1996}. Hostile and benevolent are independent flags: a speech may contain either, both, or neither. Each row pairs a hostile subcategory with its benevolent counterpart (paternalism, gender differentiation, heterosexuality).}
\label{tab:taxonomy}
\end{table}

\section{Annotation}
\label{sec:annotation}

We construct a 300-speech validation set (denoted by $\mathcal{D}_\textit{validation}$) by drawing a uniformly random sample (span all eras of our analysis in proportion to corpus density) from the 6,531 keyword-extracted speeches after codebook development and validation. Annotators (one male, one female) were presented with each speech alongside up to 5 preceding and 5 following speeches as context, mirroring the setup provided to the LLM judge. For each speech, annotators classified:

\begin{compactitem}
    \item \textbf{Relevance and Stance:} Is this speech relevant to women's political rights and representation? If relevant, classify as For, Against, or Both.
    \item \textbf{Type of Sexism:} What type of sexism, if any, is expressed by the speaker? Mark all applicable labels from Table~\ref{tab:taxonomy}; hostile and benevolent are independent flags, each with subcategory tags.
\end{compactitem}

Raters worked independently using a custom Streamlit application (to avoid era-clustering biases) that presented speeches in the randomised order. After independent annotation, raters resolved 106 cases of disagreement through discussion (62 on stance, 44 on sexism subcategories only), producing human labels for all of $\mathcal{D}_\textit{validation}$.

Two computational social science researchers with familiarity with historical political discourse conducted this annotation. Crowdsourcing was unsuitable: Hansard debates feature archaic English, complex parliamentary conventions, and lengthy rhetorical speeches that demand interpretive judgment grounded in political and social theory. Each annotator independently labelled $\mathcal{D}_\textit{validation}$ (median 890 words per speech; 95th percentile 3{,}521 words), investing  $\sim$40 hours per annotator. Given this annotation intensity, we prioritised depth and reliability over scale, producing a smaller but rigorously labelled evaluation set.

\section{Methods}

\subsection{Suffrage Stance Classification}

We classified the stance (for / against / both / irrelevant) towards women's suffrage and franchise in 6,531 speeches identified through keyword-based retrieval using Claude Sonnet 4.6 via the Anthropic API. For each target speech, we provided up to 5 preceding and 5 following speeches as context to help the model understand references and implicit arguments, similar to the human annotation setup. We selected Claude Sonnet 4.6 after comparing multiple models and prompting strategies, as described in Section \ref{subsec:val_res}. The full prompt is provided in Appendix \ref{sec:prompt}. Because many suffrage debates also touch on broader questions of women's political representation (e.g., standing for Parliament, serving on committees), our classification prompt defines relevance broadly to capture these related arguments when they arise in the context of franchise debates. 

\subsection{Sexism Classification}
We separately perform classification on sexism based on Table~\ref{tab:taxonomy}. We use the same context and settings  as suffrage classification, and the full prompt is provided in Appendix \ref{sec:prompt}. We prompt the model to decide on the presence and subtype of hostile and benevolent sexism in the target speech. Sexism classification is difficult for both humans and LLMs (see Section \ref{subsec:val_res}), so we assign higher uncertainty to our interpretation of these classification results.

\section{Results}
\subsection{Annotation and Classification Validation}
\label{subsec:val_res}

\begin{table}[t]
\centering
\footnotesize
\begin{tabular}{@{}lrrrr@{}}
\toprule
\textbf{Class} & \textbf{P} & \textbf{R} & \textbf{F1} & \textbf{$n$} \\
\midrule
For        & 0.82 & 0.79 & 0.80 & 96  \\
Against    & 0.81 & 0.70 & 0.75 & 30  \\
Both       & 0.40 & 0.15 & 0.22 & 13  \\
Irrelevant & 0.86 & 0.94 & 0.90 & 161 \\
\midrule
Macro avg.  & 0.72 & 0.65 & 0.67 & 300 \\
Weighted   & 0.81 & 0.83 & 0.82 & 300 \\
\bottomrule
\end{tabular}
\caption{Per-class stance classification metrics for Claude Sonnet 4.6 against human labels (n=300). The ``Both'' class is rare (4\% of validation) and the LLM under-detects it; for the remaining classes (For, Against, Irrelevant), the LLM achieves $F1 \geq 0.75$.}
\label{tab:per_class_stance}
\end{table}

Human--human agreement on stance was substantial (Cohen's $\kappa = 0.644$). Raters then resolved 62 cases of disagreement on stance through discussion, producing human labels for all of $\mathcal{D}_\textit{validation}$.

The speeches in $\mathcal{D}_\textit{validation}$ were also annotated for hostile and benevolent sexism. The two annotators flagged hostile sexism in 28 speeches, benevolent in 54, with 16 of those flagged for both, making up 66 unique sexist speeches overall (22\% of the sample). Claude flagged 14 hostile and 32 benevolent (39 unique). Appendix~\ref{sec:sexism_class_agreement} reports Cohen's $\kappa$: the two annotators agree with each other moderately on the binary flags, and Claude's agreement with the final human labels is higher. This is expected since the human labels are determined after the annotators resolved their disagreements, so Claude is being compared against a cleaner target than the annotators were comparing against each other. Either way, sexism annotation is genuinely hard for humans on this kind of historical material. When Claude flags a speech as sexist it is usually correct (precision 0.77--0.86), but it misses roughly half the cases the annotators would flag (recall 0.43--0.46). The per-stance sexism rates we report are therefore likely underestimates of the true rates (Refer Appendix~\ref{sec:classification_reliability}).

Table~\ref{tab:baselines} compares classification methods on $\mathcal{D}_\textit{validation}$. Standard baselines (majority class, DeBERTa-v3 zero-shot, TF-IDF with logistic regression) fall well short of the LLM. Claude Sonnet 4.6 reaches $\kappa = 0.711$ against human labels, comparable to the two annotators' agreement with each other ($\kappa = 0.644$). Per-class precision, recall, and F1 are reported in Table~\ref{tab:per_class_stance}; the well-represented classes (For, Against, Irrelevant) all achieve $F1 \geq 0.75$, while the rare ``Both'' class (only 13 cases in the 300) is under-detected.

The distributional patterns we observe also reproduce the structure predicted by the Ambivalent Sexism Inventory: hostile sexism is concentrated in speeches opposing women's rights, benevolent sexism in those that support. To rule out the possibility that the hostile/benevolent distinction simply reflects sentiment, we ran a sentiment classifier (DistilBERT fine-tuned on SST-2) on all 886 sexist speeches in our corpus (Appendix~\ref{sec:sentiment}). Within each stance category, both hostile and benevolent speeches are overwhelmingly negative in sentiment (among anti-suffrage speeches: 88\% of hostile and 86\% of benevolent; overall 87\% hostile and 77\% benevolent). Because both forms are predominantly negative-sentiment, a sentiment classifier alone cannot distinguish them, confirming that the hostile/benevolent taxonomy captures rhetorical structure beyond surface-level sentiment.

LLM-as-a-judge classifiers are known to show model-specific biases (e.g., preferring longer or more confident-sounding responses, behaving inconsistently across model families, etc.) \citep{chehbouni2025neither}. To check that our results do not reflect any single model's quirks, we also classified the $\mathcal{D}_\textit{validation}$ speeches with three additional large language models (GPT-5, Gemini 2.5 Flash, DeepSeek V3) and compared pairwise agreement between models and the human consensus labels. We additionally compare against standard non-LLM baselines: a majority-class baseline, TF-IDF with logistic regression (5-fold CV), and DeBERTa-v3 zero-shot via NLI \citep{he2023debertav3}.

\begin{table}[t]
\centering
\footnotesize
\begin{tabular}{@{}L{0.54\columnwidth}R{0.18\columnwidth}R{0.18\columnwidth}@{}}
\toprule
\textbf{Method} & \textbf{Agree} & \textbf{$\kappa$} \\
\midrule
Majority baseline & 53.7\% & 0.000 \\
DeBERTa-v3 zero-shot (NLI) & 50.3\% & 0.269 \\
TF-IDF + LogReg (5-fold CV) & 68.7\% & 0.419 \\
\addlinespace[3pt]
Human--human ($n{=}300$) & 79.3\% & 0.644 \\
\textbf{Claude Sonnet 4.6} & \textbf{83.3\%} & \textbf{0.711} \\
\bottomrule
\end{tabular}
\caption{Stance classification agreement on $\mathcal{D}_\textit{validation}$. Human labels are determined by unanimous annotator agreement and discussion-based resolution of disagreements. All baselines and Cohen's $\kappa$ are computed against these consensus human labels.}
\label{tab:baselines}
\end{table}

\begin{table}[t]
\centering
\footnotesize
\begin{tabular}{@{}lrrrrr@{}}
\toprule
& \textbf{H} & \textbf{C} & \textbf{G} & \textbf{F} & \textbf{D} \\
\midrule
Human    & --    & .711 & .692 & .533 & .658 \\
Claude   & .711  & --   & .739 & .593 & .777 \\
GPT-5    & .692  & .739 & --   & .605 & .709 \\
Gemini   & .533  & .593 & .605 & --   & .555 \\
DeepSeek & .658  & .777 & .709 & .555 & --   \\
\bottomrule
\end{tabular}
\caption{Pairwise stance agreement (Cohen's $\kappa$) on $\mathcal{D}_\textit{validation}$. H = Human consensus, C = Claude Sonnet 4.6, G = GPT-5, F = Gemini 2.5 Flash, D = DeepSeek V3. All four models received identical prompts and speech context windows.}
\label{tab:cross_llm}
\end{table}

\subsection{Stance Classification}
\label{subsec:class_res}
Of 6{,}531 extracted speeches, Sonnet 4.6 deemed 55\% irrelevant, and of the 2{,}942 relevant speeches, 74\% are for, 19\% are against, and 7\% are both. Appendix~\ref{sec:irrelevant_distribution} further discusses the spread of Irrelevant speeches over the decades. Table~\ref{tab:stance_dist} shows the distribution by era and speaker gender. Table~\ref{tab:sexism_examples} gives illustrative samples of each stance. We see an increasing trend in For stances over time, with splits chosen around the legislative milestones of 1918 and 1928.

Overall, the representation of women is viewed positively by lawmakers over the past two centuries, as one would expect given the steady progress in women's rights and freedoms both in the UK and across the world. We also see that over time, opposition to women's representation steadily drops, with a negligible amount remaining post-1950.

We observe a stark divide in stance based on the MP's gender. Female MPs are 93\% in support of women's representation, while male MPs are 70\%. This difference is statistically significant ($\chi^2 = 86.75$, $p < 0.001$). A logistic regression controlling for decade confirms that gender remains a significant predictor of supportive stance (odds ratio 2.01, $p = 0.002$), though part of the gap reflects women entering Parliament in more supportive eras (Appendix~\ref{sec:gender_era}). Female MPs remain steadfast across eras (92--97\% support), while male MP support rises from 54\% in 1870--1899 to 91\% post-1950 and converging to female-comparable levels only after questions of suffrage were settled.

While stance distribution captures \textit{whether} a speaker is supportive of women's political representation, it does not tell us \textit{how} they argue for or against. We turn to this question in the next section. 
\begin{figure*}[t]
\centering
\includegraphics[width=\textwidth]{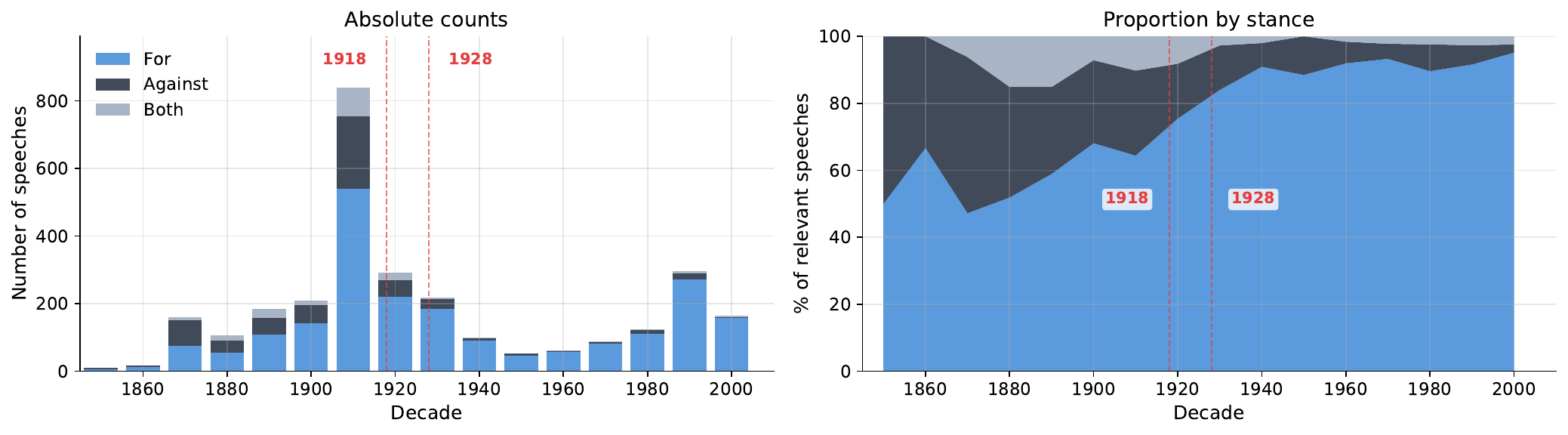}
\caption{Stance distribution by decade. Left: absolute counts. Right: percentage of relevant speeches. Dashed lines mark the 1918 and 1928 Representation of the People Acts. Opposition collapses after enfranchisement.}
\label{fig:temporal}
\end{figure*}

% Stance distribution by era and gender
 
\begin{table}[t]
\centering
\footnotesize
\begin{tabular}{@{}lrrrr@{}}
\toprule
& \textbf{n} & \textbf{For \%} & \textbf{Against \%} & \textbf{Both \%} \\
\midrule
\multicolumn{5}{@{}l}{\textit{By era}} \\
pre-1870     & 39    & 54 & 46 & 0 \\
1870--1899   & 452   & 53 & 35 & 12 \\
1900--1918   & 964   & 64 & 27 & 9 \\
1919--1928   & 374   & 75 & 15 & 10 \\
1929--1950   & 326   & 87 & 11 & 2 \\
post-1950    & 787   & 92 & 6  & 2 \\
\addlinespace[3pt]
\multicolumn{5}{@{}l}{\textit{By speaker gender}} \\
Female MPs   & 389   & 93 & 4  & 3 \\
Male MPs     & 2{,}303 & 70 & 22 & 8 \\
\midrule
\textbf{Overall} & \textbf{2{,}942} & \textbf{74} & \textbf{19} & \textbf{7} \\
\bottomrule
\end{tabular}
\caption{Stance distribution by era and speaker gender (\% of relevant speeches).}
\label{tab:stance_dist}
\end{table}

\begin{table*}[t]
\centering
\footnotesize
\begin{tabular}{@{}L{0.07\textwidth}L{0.16\textwidth}L{0.20\textwidth}L{0.51\textwidth}@{}}
\toprule
\textbf{Stance} & \textbf{Speaker} & \textbf{Classification} & \textbf{Excerpt} \\
\midrule
Against & Sir B.\ Simeon (1897)
& Hostile
& ``\ldots if ever [women] got into Parliament the end of this country would not be far off.'' \\[6pt]

Against & Mr.\ H.\ Robertson (1905)
& Hostile
& ``Women were too apt to go into enormous detail, and they could not carry on municipal business unless people confined themselves more to general principles.'' \\[6pt]

Against & Sir G. COCKERILL (1928)
& Benevolent
& ``I think it would become woman better to assume the robe of Portia and deal out evenhanded justice between the sexes, rather than to wear the garment of Shylock and clutch at the jewelled orb of power \ldots rest content to share with man, unchallenged and unchallengeable \ldots'' \\[6pt]

For & Sir G.\ Bowyer (1867)
& Benevolent
& ``\ldots it would be manifestly indecorous for them to attend the hustings \ldots voting papers \ldots would enable the sex to vote in a manner free from objection.'' \\[6pt]

For & Rev.\ I.\ Paisley (1992)
& Benevolent
& ``I believe that a woman has a tenderness that some males do not have.'' \\[6pt]

For & Mr. Charles Roberts (1913)
& Hostile
& ``\ldots dissociate those of us who are in favour of the cause of Women Suffrage from any sympathy with these suffragette outrages \ldots if the cause does not come to the success, it is just due to the hysterical, wrong-headed action of these women \ldots'' \\[6pt]

Against & Mr. Labouchere (1894)
& Hostile + Benevolent
& ``He was opposed to any woman of any sort or kind [having] a vote \ldots He was one of the advocates of the domestic angel doctrine \ldots It would be destructive of all the charms of domesticity if women were given votes.'' \\[6pt]

Both & Mr.\ Asquith (1899)
& Hostile + Benevolent
& ``\ldots it would be intolerable to have women sitting in this House \ldots [yet I believe] it is most desirable and most useful in every way to have the influence of women in municipal work.'' \\
\bottomrule
\end{tabular}
\caption{Examples of sexism in parliamentary speech, classified by the Ambivalent Sexism Inventory \citep{glick1996}. Excerpts are verbatim from Hansard; ellipses indicate omitted text or splices across the same speech.}
\label{tab:sexism_examples}
\end{table*}

\subsection{Sexism Classification}
We classify each relevant speech for hostile and benevolent sexism following \citet{glick1996}. A speech can contain hostile sexism, benevolent sexism, both, or neither; each flag has to be supported by a verbatim quote from the speech. Of 2{,}942 relevant speeches, 886 (30\%) contain at least one form of sexism: 392 are flagged hostile, 706 benevolent, and 212 are flagged for both. Table~\ref{tab:sexism_by_stance} shows the breakdown by stance.

Among the 2{,}167 \textit{for}-speeches, 21\% contain sexism, and it is almost always benevolent (81\% of the sexist subset is benevolent-only; only 11\% is hostile-only). Among the 570 \textit{against}-speeches, 54\% contain sexism, and the two forms appear in roughly equal measure: 37\% hostile-only, 19\% benevolent-only, and 44\% containing both. \textit{Both}-stance speeches (n=205) sit between: 57\% are sexist, mostly benevolent (52\% benevolent-only).

It is striking that speakers advocating \textit{for} women's political rights exhibit sexism at all, yet 462 such speeches do, and almost all of it benevolent and paternalistic. Table~\ref{tab:sexism_examples} gives illustrative excerpts. Pro-suffrage sexism typically takes the form of praise that confines. Sir George Bowyer (1867), supporting women's right to vote, argued it would be ``manifestly indecorous for them to attend the hustings,'' and proposed voting papers so ``the sex'' could participate without public exposure. A century later, the Rev.\ Ian Paisley (1992) praised women's ``tenderness that some males do not have'' while welcoming a female Speaker. Both speakers support women's participation but attribute special qualities to women as a group, which the Ambivalent Sexism Inventory classifies as complementary gender differentiation, which is praise that fits women into roles complementing rather than competing with men.

Anti-suffrage sexism is more direct, and the same speech often combines hostile and benevolent framings. Some opponents deny women's fitness for political life outright (``Women were nervous, emotional, and had very little sense of proportion,'' Mr.\ Labouchere, 1905) or appeal to divine and natural order (``man in the beginning was ordained to rule over the woman, and this is an Eternal decree,'' Earl Percy, 1873). The same speeches frequently fold in protective paternalism to justify exclusion (``the very qualities which made women the solace of our homes \ldots unfitted them for the hard fight of politics,'' Mr.\ Beresford Hope, 1872). Even as late as 1982, Sir Nicholas Bonsor described women's participation as a ``screeching, squawking attack upon the male sex.''

The mix of sexism types also shifts over time. Hostile sexism falls from 60\% of sexist speeches in 1870--1899 to 27--30\% after 1929; benevolent sexism stays at 74--83\% throughout the corpus. Appendix~\ref{sec:speaker_aggregation} further analyses findings when prolific speakers are accounted for. Overt hostility appears to become less acceptable in parliamentary discourse during the 20th century, while benevolent framing persists. This fits the observation that benevolent sexism, disguised as positive regard for women, is harder to notice and challenge than overt hostility \citep{glick1996}.

\begin{table}[t]
\centering
\footnotesize
\begin{tabular}{@{}lrrr@{}}
\toprule
& \textbf{For} & \textbf{Against} & \textbf{Both} \\
& \textit{(n=2{,}167)} & \textit{(n=570)} & \textit{(n=205)} \\
\midrule
No sexism                & 79\% & 46\% & 43\% \\
Hostile only             & 2\%  & 20\% & 8\%  \\
Benevolent only          & 17\% & 11\% & 29\% \\
Hostile and benevolent   & 2\%  & 24\% & 20\% \\
\bottomrule
\end{tabular}
\caption{Sexism categories by stance.}
\label{tab:sexism_by_stance}
\end{table}

\section{Discussion}

Our analysis reveals that the disagreement over women's political rights in Parliament was not simply a matter of differing conclusions but of fundamentally different rhetorical strategies. Supporters who exhibited sexism did so through praise that confined, attributing positive qualities to women while implicitly restricting their roles; opponents were direct in their contempt. This pattern is consistent with the theory of ambivalent sexism, which predicts that hostile and benevolent sexism serve different social functions: hostile sexism punishes women who challenge the status quo, while benevolent sexism rewards conformity to traditional roles. Our data provides naturalistic evidence for this distinction in a real institutional setting, extending what has previously been observed primarily in survey-based studies. Moreover, the temporal shift from hostile to benevolent sexism across the 20th century suggests that overt hostility becomes socially costly in parliamentary discourse before paternalistic framing does. Further, our analysis reveals a notable paradox: despite overwhelming parliamentary support for women's suffrage (74\% in favour versus 19\% opposed), suffrage legislation took decades to pass. This disparity suggests that opponents, though a parliamentary minority, wielded disproportionate rhetorical influence and mounted sustained resistance against the bills.

The persistence of benevolent sexism even among supporters of women's rights suggests that progressive policy positions do not require progressive reasoning. A speaker can advocate for women's enfranchisement while still framing women as fragile, morally pure, or in need of special accommodation. This has implications beyond historical analysis: if sexist assumptions can coexist with supportive policy positions, they may be harder to identify and challenge in contemporary discourse as well.

The struggle for suffrage that we document here is not an isolated incident. Although it has been argued that we live in the most equal time in history \citep{piketty2014capital}, rights are neither universal nor irreversible; reproductive rights in the United States were recently curtailed after fifty years when \textit{Roe v.\ Wade} was overturned. Legislative debates over group rights, whether women's suffrage in 1918, equal marriage in 2004, or ongoing deliberations in international bodies, share a common discursive structure in which supporters and opponents deploy different rhetorical strategies to argue over who deserves full participation in public life. The methods we develop here, combining LLM-based classification with established social-psychological taxonomies, are directly applicable to these settings. Any type of legislative debate corpus could potentially be analysed for the type, and not merely the direction, of discriminatory reasoning, enabling researchers to study the rhetorical structures that either buttress or challenge inequality in society. More broadly, richer LLM capabilities make historical text tractable at scale. Studies like \citet{card2022computational}, tracing 140 years of US immigration framing, and our analysis of 200 years of Hansard debate show that LLMs offer a novel lens on history. They surface latent rhetorical patterns across corpora far too large for manual close reading, letting researchers ask not just what was said, but how reasoning about contested groups shifted across centuries.

\section{Conclusion}
When Earl Percy declared in 1873 that ``man in the beginning was ordained to rule over the woman, and this is an Eternal decree,'' he was deploying a specific type of reasoning that our taxonomy classifies as hostile sexism, i.e., a contemptuous framing that asserts male authority by denigrating women. A century later, while the frequency of arguments by those \textit{against} women's representation had reduced considerably, the nature of the discourse had not. Even more strikingly, speakers \textit{for} women's political rights would still display sexist reasoning at times, but their reasoning mapped onto benevolent sexism --- paternalistic praise that idealizes women's distinctive qualities while restricting their roles. That two distinct rhetorical modes of gendered reasoning are separated across speaker stances on women's political participation, mapping onto the hostile/benevolent structure predicted by the Ambivalent Sexism Inventory, suggests that the mechanisms of reasoning behind sexist speech are remarkably stable across contexts and eras.

In this work, we extracted and enriched a British parliamentary dataset of over 6.7 million speeches from 200 years of debate and examined how discourse around women's political representation varied by stance, gender, and era. We showed that opposing sides do not simply disagree, but they reason differently about gender, and that the type of sexism deployed is as informative as its presence. We release this dataset, which adds useful speaker metadata to and reorganizes the Hansard corpus, to enable further computational social science research on historical political discourse \footnote{\url{https://github.com/Social-Insights-Lab/Two-Centuries-of-Sexism}}.

\section{Acknowledgements}

We thank Praharshita Kaithepalli, Shaista Syeda, Nandhini Shankar Lakshman, and Niranjana Deshpande for conducting preliminary investigations that informed this work. We also thank Dr. Chritopher M. Homan and Dr. Evan Selinger for their support and feedback as part of Mandira Sawkar’s thesis committee. KhudaBukhsh was partly supported by a gift from Lenovo.

\section{Limitations}

Our analysis has several limitations. We classify sexism using a single primary LLM judge (Claude Sonnet 4.6), though cross-model validation with GPT-5, Gemini 2.5 Flash, and DeepSeek V3 suggests the classifications reflect textual signal rather than model-specific artefacts. The keyword-based retrieval of suffrage speeches cuts both ways: our filter can both miss speeches that discuss women's political rights without using our search terms, and pick up speeches where the keyword refers to something else (e.g.\ ``franchise'' in a debate about male voting rights). The LLM relevance classifier filters out the second kind (Section~\ref{subsec:class_res}), but the first kind sets an upper bound on what we can find. We analyse only the House of Commons. The correlation between sexism type and stance may partly reflect detection difficulty: hostile sexism is more obvious and easier for the LLM to identify than benevolent sexism, which can be embedded in superficially positive language. Our sentiment confound analysis (Appendix~\ref{sec:sentiment}) partially addresses this concern, showing both forms are predominantly negative-sentiment, so a sentiment classifier alone cannot distinguish them although a residual asymmetry in detection difficulty is possible. Any study on binary gender bias runs the risk of oversimplifying gender. Also, parliamentary speech should be understood as a form of institutional discourse shaped by norms of formality, persuasion, and audience expectations. Speakers may frame arguments in ways that align with what is considered rhetorically acceptable within Parliament, rather than directly expressing private beliefs.

\section{Ethical Considerations}

We reproduce verbatim sexist quotes from parliamentary speeches for the purpose of scholarly analysis, not endorsement. The Hansard data is publicly available government records.  We acknowledge that gender lies on a spectrum \citep{banerjee2024gender}, and our binary classification of speaker gender reflects the categories available in the historical parliamentary records rather than the full range of gender identities. Following \citet{banerjee2024gender}, we conducted a noise induction robustness check (Appendix \ref{sec:noise}): in each of 1,000 iterations, we randomly flipped 3\% of speaker gender labels (M to F or F to M) and recomputed the gender-stance analysis. The gender difference remained statistically significant ($p < 0.05$) in all 1{,}000 iterations (Appendix~\ref{sec:noise}, Table~\ref{tab:noise}), confirming our conclusions are robust to potential misgendering in the speaker-matching pipeline. The framework proposed is also primarily descriptive, aiming to categorize recurring patterns in how gender bias is expressed rather than to provide causal explanations for why individuals in specific historical contexts held particular beliefs. As such, they do not attempt to model historically contingent motivations (e.g., religious or cosmological justifications), but instead offer a structured vocabulary for analysing the form of sexist reasoning across contexts. Additionally, while these findings might be novel within the context of NLP and large-scale computational analysis, we do not claim that they are necessarily surprising from the perspective of feminist history or sociology. Scholars in these fields have long examined the role of both overt hostility and paternalistic reasoning in shaping women’s rights movements. Our contribution lies not in overturning these insights, but in providing large-scale, quantitative evidence of these patterns in naturalistic parliamentary discourse.

\section*{Use of AI Assistants}

We used Claude Code and Claude.ai (Anthropic) for code development, experimental infrastructure, and polishing prose during the writing process. We used OpenAI's Deep Research feature to assist us with literature search. Claude Sonnet 4.6 was used as the primary LLM judge for stance and sexism classification, which is extensively documented in our methods and validation sections. GPT-5, Gemini 2.5 Flash, and Deepseek V3 were used as secondary LLM judges and similarly documented. All scientific decisions, interpretations, and core arguments were made by the authors.

\bibliography{custom}

\begin{thebibliography}{42}
\providecommand{\natexlab}[1]{#1}

\bibitem[{Abercrombie and Batista-Navarro(2020{\natexlab{a}})}]{abercrombie-batista-navarro-2020-parlvote}
Gavin Abercrombie and Riza Batista-Navarro. 2020{\natexlab{a}}.
\newblock \href {https://aclanthology.org/2020.lrec-1.624/} {{P}arl{V}ote: A corpus for sentiment analysis of political debates}.
\newblock In \emph{Proceedings of the Twelfth Language Resources and Evaluation Conference}, pages 5073--5078, Marseille, France. European Language Resources Association.

\bibitem[{Abercrombie and Batista-Navarro(2020{\natexlab{b}})}]{Abercrombie2020-lit-rev}
Gavin Abercrombie and Riza Batista-Navarro. 2020{\natexlab{b}}.
\newblock \href {https://doi.org/10.1007/s42001-019-00060-w} {Sentiment and position-taking analysis of parliamentary debates: a systematic literature review}.
\newblock \emph{Journal of Computational Social Science}, 3(1):245--270.

\bibitem[{Abercrombie and Batista-Navarro(2022)}]{abercrombie-batista-navarro-2022-policy}
Gavin Abercrombie and Riza Batista-Navarro. 2022.
\newblock \href {https://doi.org/10.3384/nejlt.2000-1533.2022.3454} {Policy-focused stance detection in parliamentary debate speeches}.
\newblock \emph{Northern European Journal of Language Technology}, 8.

\bibitem[{Abercrombie and Batista-Navarro(2018)}]{Abercrombie2018}
Gavin Abercrombie and Riza~Theresa Batista-Navarro. 2018.
\newblock \href {http://lrec-conf.org/workshops/lrec2018/W2/pdf/9_W2.pdf} {A sentiment-labelled corpus of hansard parliamentary debate speeches}.
\newblock In \emph{Proceedings of the Eleventh International Conference on Language Resources and Evaluation (LREC 2018)}, Paris, France. European Language Resources Association (ELRA).

\bibitem[{Alexander and Struan(2022)}]{Alexander-etal}
Marc Alexander and Andrew Struan. 2022.
\newblock \href {https://doi.org/10.1075/ijcl.22016.ale} {“in barbarous times and in uncivilized countries”}.
\newblock \emph{International Journal of Corpus Linguistics}, 27(4):480--505.

\bibitem[{Altameemi et~al.(2025)Altameemi, Altamimi, Alkhalil, Uliyan, and Mansour}]{Altameemi-etal}
Yaser Altameemi, Mohammed Altamimi, Adel Alkhalil, Diaa Uliyan, and Romany~F. Mansour. 2025.
\newblock \href {https://doi.org/10.3389/frai.2025.1654496} {Text summarization method of argumentative discourse by combining the bert-transformer model}.
\newblock \emph{Frontiers in Artificial Intelligence}, Volume 8 - 2025.

\bibitem[{Banerjee et~al.(2024)Banerjee, Dutta, Datta, and KhudaBukhsh}]{banerjee2024gender}
Somonnoy Banerjee, Sujan Dutta, Soumyajit Datta, and Ashiqur~R. KhudaBukhsh. 2024.
\newblock \href {https://arxiv.org/abs/2409.12194} {Gender representation and bias in indian civil service mock interviews}.

\bibitem[{Blaxill and Beelen(2016)}]{Blaxill-etal}
Luke Blaxill and Kaspar Beelen. 2016.
\newblock \href {https://doi.org/10.1093/tcbh/hww028} {A feminized language of democracy? the representation of women at westminster since 1945}.
\newblock \emph{Twentieth Century British History}, 27(3):412--449.

\bibitem[{Card et~al.(2022)Card, Chang, Becker, Mendelsohn, Voigt, Boustan, Abramitzky, and Jurafsky}]{card2022computational}
Dallas Card, Serina Chang, Chris Becker, Julia Mendelsohn, Rob Voigt, Leah Boustan, Ran Abramitzky, and Dan Jurafsky. 2022.
\newblock \href {https://doi.org/10.1073/pnas.2120510119} {Computational analysis of 140 years of us political speeches reveals more positive but increasingly polarized framing of immigration}.
\newblock \emph{Proceedings of the National Academy of Sciences}, 119(31):e2120510119.

\bibitem[{Chehbouni et~al.(2025)Chehbouni, Haddou, Cheung, and Farnadi}]{chehbouni2025neither}
Khaoula Chehbouni, Mohammed Haddou, Jackie Chi~Kit Cheung, and Golnoosh Farnadi. 2025.
\newblock \href {https://neurips.cc/virtual/2025/loc/san-diego/poster/121914} {Neither valid nor reliable? investigating the use of {LLMs} as judges}.
\newblock In \emph{Advances in Neural Information Processing Systems}.

\bibitem[{Ding et~al.(2025)Ding, Zhao, Jia, Wang, Qian, Chen, and Yue}]{ding-etal-2025-gender}
YiTian Ding, Jinman Zhao, Chen Jia, Yining Wang, Zifan Qian, Weizhe Chen, and Xingyu Yue. 2025.
\newblock \href {https://doi.org/10.18653/v1/2025.trustnlp-main.36} {Gender bias in large language models across multiple languages: A case study of {C}hat{GPT}}.
\newblock In \emph{Proceedings of the 5th Workshop on Trustworthy NLP (TrustNLP 2025)}, pages 552--579. Association for Computational Linguistics.

\bibitem[{Duthie et~al.(2016)Duthie, Budzynska, and Reed}]{Duthie-etal}
Rory Duthie, Katarzyna Budzynska, and Chris Reed. 2016.
\newblock \href {https://doi.org/10.3233/978-1-61499-686-6-299} {Mining ethos in political debate}.

\bibitem[{Dutta et~al.(2022)Dutta, Li, Nagin, and KhudaBukhsh}]{dutta2022murder}
Sujan Dutta, Beibei Li, Daniel~S. Nagin, and Ashiqur~R. KhudaBukhsh. 2022.
\newblock \href {https://doi.org/10.24963/ijcai.2022/702} {A murder and protests, the capitol riot, and the chauvin trial: Estimating disparate news media stance}.
\newblock In \emph{Proceedings of the Thirty-First International Joint Conference on Artificial Intelligence, {IJCAI-22}}, pages 5059--5065. International Joint Conferences on Artificial Intelligence Organization.
\newblock AI for Good.

\bibitem[{Glick and Fiske(1996)}]{glick1996}
Peter Glick and \{Susan T.\} Fiske. 1996.
\newblock \href {https://doi.org/10.1037/0022-3514.70.3.491} {The ambivalent sexism inventory: Differentiating hostile and benevolent sexism}.
\newblock \emph{Journal of personality and social psychology}, 70(3):491--512.

\bibitem[{Guest et~al.(2021)Guest, Vidgen, Mittos, Sastry, Tyson, and Margetts}]{guest-etal-2021-expert}
Ella Guest, Bertie Vidgen, Alexandros Mittos, Nishanth Sastry, Gareth Tyson, and Helen Margetts. 2021.
\newblock \href {https://doi.org/10.18653/v1/2021.eacl-main.114} {An expert annotated dataset for the detection of online misogyny}.
\newblock In \emph{Proceedings of the 16th Conference of the European Chapter of the Association for Computational Linguistics: Main Volume}, pages 1336--1350, Online. Association for Computational Linguistics.

\bibitem[{Halterman et~al.(2021)Halterman, Keith, Sarwar, and O{'}Connor}]{halterman2021corpus}
Andrew Halterman, Katherine~A. Keith, Sheikh Sarwar, and Brendan O{'}Connor. 2021.
\newblock \href {https://doi.org/10.18653/v1/2021.findings-acl.371} {Corpus-level evaluation for event {QA}: The {I}ndia{P}olice{E}vents corpus covering the 2002 {G}ujarat violence}.
\newblock In \emph{Findings of the Association for Computational Linguistics: ACL-IJCNLP 2021}, pages 4240--4253, Online. Association for Computational Linguistics.

\bibitem[{Hargrave and Blumenau(2022)}]{Hargrave_Blumenau_2022}
Lotte Hargrave and Jack Blumenau. 2022.
\newblock \href {https://doi.org/10.1017/S0007123421000648} {No longer conforming to stereotypes? gender, political style and parliamentary debate in the uk}.
\newblock \emph{British Journal of Political Science}, 52(4):1584–1601.

\bibitem[{He et~al.(2023)He, Gao, and Chen}]{he2023debertav3}
Pengcheng He, Jianfeng Gao, and Weizhu Chen. 2023.
\newblock \href {https://arxiv.org/abs/2111.09543} {Debertav3: Improving deberta using electra-style pre-training with gradient-disentangled embedding sharing}.

\bibitem[{Holton(1996)}]{holton2002suffrage}
Sandra Holton. 1996.
\newblock \href {https://doi.org/10.4324/9780203427569} {\emph{Suffrage Days: stories from the women's suffrage movement}}.
\newblock Routledge.

\bibitem[{Jha and Mamidi(2017)}]{jha-mamidi-2017-compliment}
Akshita Jha and Radhika Mamidi. 2017.
\newblock \href {https://doi.org/10.18653/v1/W17-2902} {When does a compliment become sexist? analysis and classification of ambivalent sexism using twitter data}.
\newblock In \emph{Proceedings of the Second Workshop on {NLP} and Computational Social Science}, pages 7--16, Vancouver, Canada. Association for Computational Linguistics.

\bibitem[{Khadilkar et~al.(2022)Khadilkar, KhudaBukhsh, and Mitchell}]{khadilkar2022gender}
Kunal Khadilkar, Ashiqur~R KhudaBukhsh, and Tom~M Mitchell. 2022.
\newblock Gender bias, social bias, and representation in bollywood and hollywood.
\newblock \emph{Patterns}, 3(2).

\bibitem[{Kirk et~al.(2023)Kirk, Yin, Vidgen, and R{\"o}ttger}]{kirk-etal-2023-semeval}
Hannah Kirk, Wenjie Yin, Bertie Vidgen, and Paul R{\"o}ttger. 2023.
\newblock \href {https://doi.org/10.18653/v1/2023.semeval-1.305} {{S}em{E}val-2023 task 10: Explainable detection of online sexism}.
\newblock In \emph{Proceedings of the 17th International Workshop on Semantic Evaluation (SemEval-2023)}, pages 2193--2210, Toronto, Canada. Association for Computational Linguistics.

\bibitem[{Kotek et~al.(2023)Kotek, Dockum, and Sun}]{Kotek_etal}
Hadas Kotek, Rikker Dockum, and David Sun. 2023.
\newblock \href {https://doi.org/10.1145/3582269.3615599} {Gender bias and stereotypes in large language models}.
\newblock In \emph{Proceedings of The ACM Collective Intelligence Conference}, CI '23, page 12–24, New York, NY, USA. Association for Computing Machinery.

\bibitem[{Kumar et~al.(2020)Kumar, Bhotia, Kumar, and Chakraborty}]{kumar2020nurse}
Vaibhav Kumar, Tenzin~Singhay Bhotia, Vaibhav Kumar, and Tanmoy Chakraborty. 2020.
\newblock Nurse is closer to woman than surgeon? mitigating gender-biased proximities in word embeddings.
\newblock \emph{Transactions of the Association for Computational Linguistics}, 8:486--503.

\bibitem[{Li et~al.(2020)Li, Durmus, and Cardie}]{li-etal-2020-exploring-role}
Jialu Li, Esin Durmus, and Claire Cardie. 2020.
\newblock \href {https://doi.org/10.18653/v1/2020.emnlp-main.716} {Exploring the role of argument structure in online debate persuasion}.
\newblock In \emph{Proceedings of the 2020 Conference on Empirical Methods in Natural Language Processing (EMNLP)}, pages 8905--8912, Online. Association for Computational Linguistics.

\bibitem[{Lippi and Torroni(2016)}]{Lippi_Torroni_2016}
Marco Lippi and Paolo Torroni. 2016.
\newblock \href {https://doi.org/10.1609/aaai.v30i1.10384} {Argument mining from speech: Detecting claims in political debates}.
\newblock \emph{Proceedings of the AAAI Conference on Artificial Intelligence}, 30(1).

\bibitem[{Mirza et~al.(2025)Mirza, Jafari, Ozcinar, and Anbarjafari}]{info16050358}
Imran Mirza, Akbar~Anbar Jafari, Cagri Ozcinar, and Gholamreza Anbarjafari. 2025.
\newblock \href {https://doi.org/10.3390/info16050358} {Quantifying gender bias in large language models using information-theoretic and statistical analysis}.
\newblock \emph{Information}, 16(5).

\bibitem[{Mochtak et~al.(2025)Mochtak, Rupnik, Kuzman, and Ljubešić}]{Mochtak31122025}
Michal Mochtak, Peter Rupnik, Taja Kuzman, and Nikola Ljubešić. 2025.
\newblock \href {https://doi.org/10.1080/2474736X.2025.2508377} {Parlasent: mapping sentiment in political discourse with large language models}.
\newblock \emph{Political Research Exchange}, 7(1):2508377.

\bibitem[{Ogg(1918)}]{representation1918act_journal}
Frederic~A. Ogg. 1918.
\newblock \href {http://www.jstor.org/stable/1946102} {The british representation of the people act}.
\newblock \emph{The American Political Science Review}, 12(3):498--503.

\bibitem[{Parliament(1918)}]{representation1918act_parliament}
UK~Parliament. 1918.
\newblock \href {https://www.parliament.uk/about/living-heritage/transformingsociety/electionsvoting/womenvote/case-study-the-right-to-vote/the-right-to-vote/birmingham-and-the-equal-franchise/1918-representation-of-the-people-act/} {1918 representation of the people act}.

\bibitem[{Parliament(1928)}]{franchise1928act_parliament}
UK~Parliament. 1928.
\newblock \href {https://www.parliament.uk/about/living-heritage/transformingsociety/electionsvoting/womenvote/case-study-the-right-to-vote/the-right-to-vote/birmingham-and-the-equal-franchise/1928-equal-franchise-act/} {1928 equal franchise act}.

\bibitem[{Parliament(2018)}]{HistoricHansard}
UK~Parliament. 2018.
\newblock \href {https://api.parliament.uk/historic-hansard/index.html} {Historic hansard 1803-2005}.

\bibitem[{Piketty and Goldhammer(2014)}]{piketty2014capital}
Thomas Piketty and Arthur Goldhammer. 2014.
\newblock \href {http://www.jstor.org/stable/j.ctt6wpqbc} {\emph{Capital in the Twenty-First Century}}.
\newblock Harvard University Press.

\bibitem[{Raiber and Spierings(2022)}]{Anagnosticapproachtogender}
Klara Raiber and Niels Spierings. 2022.
\newblock \href {https://doi.org/10.1332/251510821X16539489608628} {An agnostic approach to gender patterns in parliamentary speech: a question of representation by topic and style}.
\newblock \emph{European Journal of Politics and Gender}, 5(3):361 -- 381.

\bibitem[{Slapin and Proksch(2008)}]{slapin-etal}
Jonathan~B. Slapin and Sven-Oliver Proksch. 2008.
\newblock \href {https://doi.org/10.1111/j.1540-5907.2008.00338.x} {A scaling model for estimating time-series party positions from texts}.
\newblock \emph{American Journal of Political Science}, 52(3):705--722.

\bibitem[{Soriano-Jiménez(2024)}]{Soriano-Jiménez_2024}
Carlos Soriano-Jiménez. 2024.
\newblock \href {https://doi.org/10.28914/Atlantis-2024-46.2.03} {Language change in british parliamentary discourse: A corpus-based study of power and authority markers, 1930-2005}.
\newblock \emph{Atlantis. Journal of the Spanish Association for Anglo-American Studies}, 46(2):49–68.

\bibitem[{Taylor(2022)}]{Taylor_2022}
Miles Taylor. 2022.
\newblock \href {https://doi.org/10.1017/S0018246X2100073X} {Parliamentary representation in modern britain: Past, present, and future}.
\newblock \emph{The Historical Journal}, 65(4):1145–1173.

\bibitem[{Wikipedia(2026{\natexlab{a}})}]{female-lords}
Wikipedia. 2026{\natexlab{a}}.
\newblock \href {https://en.wikipedia.org/wiki/List_of_female_members_of_the_House_of_Lords} {List of female members of the house of lords}.

\bibitem[{Wikipedia(2026{\natexlab{b}})}]{female-commons}
Wikipedia. 2026{\natexlab{b}}.
\newblock \href {https://en.wikipedia.org/wiki/List_of_female_members_of_the_Parliament_of_the_United_Kingdom} {List of female members of the parliament of the united kingdom}.

\bibitem[{Wingerden(1999)}]{van2016women}
Sophia~A. Wingerden. 1999.
\newblock \href {https://doi.org/10.1007/978-1-349-27493-2} {\emph{The women's suffrage movement in Britain, 1866-1928}}.
\newblock Palgrave Macmillan London.

\bibitem[{Yılmaz and Mccallion(2025)}]{Yılmaz06102025}
Atakan Yılmaz and Anne-Marie Mccallion. 2025.
\newblock \href {https://doi.org/10.1080/09589236.2025.2566669} {High politics, low visibility: a computational analysis of gender discourse in eu institutions}.
\newblock \emph{Journal of Gender Studies}, 0(0):1--28.

\bibitem[{Zeinert et~al.(2021)Zeinert, Inie, and Derczynski}]{zeinert-etal-2021-annotating}
Philine Zeinert, Nanna Inie, and Leon Derczynski. 2021.
\newblock \href {https://doi.org/10.18653/v1/2021.acl-long.247} {Annotating online misogyny}.
\newblock In \emph{Proceedings of the 59th Annual Meeting of the Association for Computational Linguistics and the 11th International Joint Conference on Natural Language Processing (Volume 1: Long Papers)}, pages 3181--3197, Online. Association for Computational Linguistics.

\end{thebibliography}

\appendix

\section{Suffrage Speech Extraction Keywords}
\label{sec:keywords}

We identified suffrage-related speeches using a two-tier keyword search applied to the full text of each speech, following prior work on corpus construction via keyword filtering \citep{halterman2021corpus, dutta2022murder}. All matching was case-insensitive.

\paragraph{Tier 1 (HIGH confidence, $n{=}2{,}725$):} Speeches containing explicit suffrage terms: \textit{women('s/woman) suffrage, female suffrage, suffrage + women, votes for women, suffragette(s), suffragist(s), enfranchise + women, women + enfranchise, equal franchise, representation of the people + women, sex disqualification, women('s) social (and) political union}.

\paragraph{Tier 2 (MEDIUM confidence, $n{=}3{,}806$):} Speeches not matching Tier 1 but containing \textit{women} or \textit{female} within 25 words of a voting-related term: \textit{vote, voting, voter(s), electoral, electorate, franchise, enfranchise, representation}.

\noindent \\Both tiers are high-recall by design; false positives (e.g., speeches where ``franchise'' refers to male voting rights) are filtered downstream by the LLM classifier, which marks 55\% of extracted speeches as irrelevant to women's political rights. Our keyword search doesn't broadly probe for non-suffrage-related women's rights speeches to keep our corpus focussed and analysis tractable from a qualitative standpoint.

However, some relevant speeches could be missed by the keyword-retrieval technique if the vocabulary diverges. For example, below is an excerpt from a speech missed by the pipeline:
\begin{quote}
\small
\textit{
    Mr. Hopkins, 1920
    \\``... I am sure the hon. Member for Plymouth will not disagree with me in this, that it is of great importance that we should legislate for the care of babies, and later on perhaps, an agitation may be got up to give the babies the vote, and we shall have a procession of perambulators going to the poll. I have no doubt the decision of the baby voters, influenced by the guile of the nurses, may be as valuable as some of the votes that are given nowadays. If this Measure had been proposed after a considerable inter-vat, when no doubt other reforms will be required, and when perhaps there will be time to take to pieces and re-examine our electoral machinery, I should have great pleasure in supporting it, I regard it to-day as premature, and I shall vote against it.'' 
}
\end{quote}

The above speech made by Mr. Hopkins in 1920 during a debate to extend the Representation of the People Bill to all women aged 21 and above (leading to the Equal Franchise Act) was not caught through keyword filtering since the word "\textit{vote}" is not within 25 words of of "\textit{women}". 

On the other hand, false positives may also occur as in the following examples:
\begin{quote}
\small
\textit{
    Prime Minister, Mr. Asquith, 1912
    \\``... As regards what the right hon. Gentleman said with reference to the reopening of the Debate on the General Navy Estimates, I think we can arrange for that before Easter. Tuesday, the 26th, is the day set aside for the Committee stage of the Consolidated Fund Bill; ... With regard to the Women's Suffrage Bill, the right hon. Gentleman, so far as I am concerned, is pushing an open door. There is some pressure from the other side. I quite agree with the right hon. Gentleman it is a necessity we should have a careful consideration of the principles of the Coal Mines Bill...''
}
\end{quote}

\begin{quote}
\small
\textit{
    Sir John Bourke, 1831
    \\``... Certain it was, that the present Bill had lost none of the supporters of the former Bill, and he believed it would be found that it had gained the votes of some of the opponents of the former measure ... he must take leave to express his dissent from the course which the Government had thought proper to pursue with respect to the Representation of Ireland. As one of the members for an Irish county, he felt bound to say, that the people of that country required a much larger number of Representatives, than it appeared was to be allotted to them; and, therefore, he now gave notice that it was his intention to move, that the Scotch and Irish Bills be brought in and read a first and second time, before they went into any Committee on the English Reform Bill...'' 
}
\end{quote}

In the above examples, the Women's Suffrage Bill is only passing mention during a discussion by the then Prime Minister, Mr. Asquith. The second quote of Sir John Burke is on the topic of Irish Representation rather than the representation of women. These samples captured by keywords are therefore \textit{Irrelevant} and marked as such by annotators.

Future work could expand the keyword list to include additional terms, more robust techniques and scale up the analysis with more computational resources.

\section{Hansard Dataset Enrichment: Speaker Gender}
\label{sec:gender_matching}
The Historical Hansard data does not include speaker gender. We developed a multi-step pipeline to infer it. We first aggregated British MP records from EveryPolitician \footnote{\url{https://everypolitician.org/}} and MySociety \footnote{\url{https://github.com/mysociety}}, merging them into a unified dataset. EveryPolitician provided explicit gender labels for a subset of MPs. For the remainder, we applied a cascading inference strategy. Honorific titles such as \textit{Mr, Ms, Countess, Duke}, etc. directly indicated gender. MPs with WikiData IDs available through MySociety's dataset were resolved via WikiData queries. We then crawled Wikipedia's MP lists \citep{female-commons, female-lords} and matched entries to the merged dataset. All MPs serving before 1918 were labelled male, as women were barred from standing for Parliament until that year. The remaining MPs were classified using the Gender Guesser library, which achieved 97\% accuracy on our data.

The resulting gendered MP dataset was linked to Hansard speakers through a combination of exact and fuzzy matching, constrained by date ranges and parliamentary titles. This pipeline achieved 89.3\% coverage of House of Commons speakers. The House of Lords proved substantially harder due to mutable peerage titles; a single individual may appear under multiple titles across time (e.g.\textit{Lord Devonshire} and \textit{Duke of Easingwold}), making reliable matching infeasible. Coverage in the House of Lords reached only 1.2\%; we therefore restricted our analysis to the House of Commons.

\section{Classification Prompt}
\label{sec:prompt}

We classify each speech in two passes. Pass~1 classifies stance; Pass~2, run only on speeches with a non-irrelevant stance, classifies sexism. Each speech (TARGET) is presented alongside up to 5 preceding and 5 following speeches (CONTEXT). Both prompts are forced through strict tool-use to guarantee schema-valid JSON output.

\subsection*{Pass 1: Stance}
\begin{quote}
\small
You are classifying parliamentary speeches from the Hansard corpus (UK Parliament, 1803--2005) for their stance on women's political rights and representation.

\textbf{Scope.} The scope of ``women's political rights'' follows the suffrage movement's framing of full civic personhood: the right to vote, the right to stand for and serve in public office, the structures of political representation, and arguments about women's fitness for political participation. We treat as in-scope any policy in which women are explicitly singled out as a class for distinctive rights, exemptions, or duties of civic participation. Health, childcare, education, or welfare policy that incidentally affects women without addressing their distinctive civic status is out of scope.

\textbf{Use of context.} Use the CONTEXT (preceding and following speech turns) to resolve references to ``the Bill'' or ``this question'', to identify arguments the speaker is responding to, and to catch irony or implicit references. Classify only the TARGET; stances expressed only in the CONTEXT do not count toward the TARGET's stance.

\textbf{Labels.} ``for'' (supports women's political rights or representation), ``against'' (opposes), ``both'' (substantive arguments on both sides --- e.g.\ supporting rights for one subset of women while opposing for another, or supporting voting while opposing office), ``irrelevant''. Procedural objection is not substantive opposition (stays ``for''); grudging acknowledgement is not endorsement (stays ``against'').
\end{quote}

\subsection*{Pass 2: Sexism (multi-label, run only on relevant speeches)}
\begin{quote}
\small
You are classifying whether a parliamentary speech contains sexism toward women under Ambivalent Sexism Inventory \citep{glick1996}. Hostile and benevolent sexism are two independent flags; a single speech may contain either, both, or neither.

\textbf{Hostile sexism} degrades, blames, or seeks to control women. Sub-types: \textit{dominative\_paternalism} (women are incompetent and require male control); \textit{competitive\_gender\_differentiation} (men are more competent in high-status domains); \textit{heterosexual\_hostility} (women's sexuality is framed as manipulative or threatening).

\textbf{Benevolent sexism} essentialises women, attributing admired traits (purity, nurturance, morality, delicacy) to women as a class, typically expressed in chivalrous or protective terms, and uses that essentialisation to justify restricting their roles. Sub-types: \textit{protective\_paternalism} (men should care for, shield, decide for, sacrifice for women); \textit{complementary\_gender\_differentiation} (women possess special purity, moral sensitivity, or tenderness that men lack); \textit{heterosexual\_intimacy} (men are framed as incomplete without women as romantic partners).

\textbf{Procedure.} Decide the hostile and benevolent binary flags based on the definitions, and list the subcategories that apply. If a binary flag is true, the corresponding subcategory list must contain at least one entry; if no subcategory fits, do not mark the binary flag. Each binary flag must be supported by a verbatim quote from the TARGET text. Use the CONTEXT only to disambiguate references; classify only the TARGET.
\end{quote}

\subsection*{Output Format}
\begin{quote}
\small
Both passes use strict tool-use, returning JSON validated against fixed schemas. Pass~1 returns \texttt{\{stance, rationale, confidence\}}. Pass~2 returns \texttt{\{hostile, hostile\_subcategories, hostile\_quote, benevolent, benevolent\_subcategories, benevolent\_quote\}}.
\end{quote}
\section{Cross-model agreement}
 \label{sec:errorbounds} 
To check how robust the classifications are across model families, we ran the two-pass classification through three additional LLMs accessed via OpenRouter (GPT-5, Gemini 2.5 Flash, DeepSeek V3) on $\mathcal{D}_\textit{validation}$. Table~\ref{tab:cross_llm_full} reports stance and sexism agreement with human labels. All four models received identical prompts. One practical note: GPT-5 is a reasoning model whose hidden reasoning tokens count against the response budget; \texttt{max\_tokens} must be set generously (we used 16{,}000) to avoid silent truncation of the tool-call response.

\begin{table}[h]
\centering
\footnotesize
\begin{tabular}{@{}lrrr@{}}
\toprule
\textbf{Model} & \textbf{Stance $\kappa$} & \textbf{Hostile $\kappa$} & \textbf{Benevolent $\kappa$} \\
\midrule
Claude Sonnet 4.6 & .711 & .543 & .463 \\
GPT-5              & .692 & .512 & .471 \\
DeepSeek V3        & .658 & .639 & .246 \\
Gemini 2.5 Flash   & .533 & .416 & .381 \\
\bottomrule
\end{tabular}
\caption{Agreement with human labels (Cohen's $\kappa$) on $\mathcal{D}_\textit{validation}$. Pairwise model--model kappa appears in Table~\ref{tab:cross_llm}.}
\label{tab:cross_llm_full}
\end{table}

\section{Sentiment Confound Analysis}
\label{sec:sentiment}

To check that the hostile/benevolent distinction is not just sentiment in disguise, we ran a sentiment classifier (DistilBERT fine-tuned on SST-2) on all 886 sexist speeches in the corpus. Table~\ref{tab:sentiment} reports the percentage classified as negative by sexism type and stance. Both hostile and benevolent speeches are overwhelmingly negative-sentiment across every stance category --- so a sentiment classifier alone cannot tell them apart.

\begin{table}[h]
\centering
\footnotesize
\begin{tabular}{@{}lrr@{}}
\toprule
\textbf{Stance} & \textbf{Hostile \% neg} & \textbf{Benevolent \% neg} \\
\midrule
Against & 88\% & 86\% \\
For     & 85\% & 73\% \\
Both    & 84\% & 80\% \\
\addlinespace[3pt]
All     & 87\% & 77\% \\
\bottomrule
\end{tabular}
\caption{Percentage of sexist speeches classified as negative-sentiment by DistilBERT (SST-2), broken down by stance and sexism type.}
\label{tab:sentiment}
\end{table}

\section{Gender-Era Confound Analysis}
\label{sec:gender_era}

Since women entered Parliament only in 1919, the raw gender gap in stance (93\% vs 70\%) conflates gender effects with temporal effects. We fit a logistic regression predicting supportive stance from speaker gender and decade as shown in Table \ref{tab:logreg}.

\begin{table}[h]
\centering
\footnotesize
\begin{tabular}{@{}lrrr@{}}
\toprule
\textbf{Predictor} & \textbf{Odds Ratio} & \textbf{$p$} \\
\midrule
Female (vs Male) & 2.01 & 0.002 \\
Decade & 1.21 & $<$0.001 \\
\bottomrule
\end{tabular}
\caption{Logistic regression predicting supportive stance (For vs not-For) from gender and decade.}
\label{tab:logreg}
\end{table}

Gender remains a significant predictor after controlling for era (OR = 2.01, $p = 0.002$), though the effect is smaller than the raw percentages suggest.

\section{Sexism Classification Agreement}
\label{sec:sexism_class_agreement}

Table \ref{tab:sexism_validation} shows inter-annotator and LLM-and-human label agreement. \textbf{Inter--annotator} compares the two annotators' independent labels before they resolved disagreements. The column \textbf{Claude} compares Claude to the final human labels.

\begin{table}[h]
\centering
\footnotesize
\begin{tabular}{@{}lcc@{}}
\toprule
                  & \textbf{Inter--annotator $\kappa$} & \textbf{Claude} \\
\midrule
Hostile sexism    & 0.302 & 0.543 \\
Benevolent sexism & 0.270 & 0.463 \\
Any sexism        & 0.338 & 0.488 \\
\bottomrule
\end{tabular}
\caption{Cohen's $\kappa$ for sexism flagging on $\mathcal{D}_\textit{validation}$.}
\label{tab:sexism_validation}
\end{table}

\section{Noise Induction Robustness Check}
\label{sec:noise}

Following \citet{banerjee2024gender}, we tested whether our gender-based findings are robust to potential misgendering in the speaker-matching pipeline. In each of 1{,}000 iterations, we randomly flipped 3\% of speaker gender labels (80 of 2{,}692 gendered relevant speeches) and recomputed the gender-stance analysis (Table~\ref{tab:noise}).

\begin{table}[h]
\centering
\footnotesize
\begin{tabular}{@{}lr@{}}
\toprule
\textbf{Metric} & \textbf{Value} \\
\midrule
Baseline male for \% & 70.1\% \\
Baseline female for \% & 92.8\% \\
Baseline $\chi^2$ & 86.75 \\
\addlinespace[3pt]
Noisy male for \% (mean) & 70.2\% \\
Noisy female for \% (mean) & 89.3\% \\
Noisy gap (mean) & 19.1pp \\
Noisy $\chi^2$ (mean) & 68.99 \\
Significant ($p < 0.05$) & 1{,}000/1{,}000 \\
\bottomrule
\end{tabular}
\caption{Noise induction robustness check. The gender difference in stance remains significant in all 1{,}000 iterations of 3\% label noise.}
\label{tab:noise}
\end{table}

The gender difference remained statistically significant ($p < 0.05$) in all 1{,}000 iterations, confirming robustness to potential misgendering.

\section{Samples of Disagreement}
\label{sec:disagreement_samples}

\subsection{Human--Human Disagreement}

The annotators labelled $\mathcal{D}_\textit{validation}$ for stance and sexism (Cohen’s $\kappa = 0.644$, 79.3\% agreement) of which they disagreed on 106 cases. They resolved these upon discussion. Below are some examples of disagreements between annotators.

\begin{itemize}
    \item \textbf{Annotator 1: Both, Hostile
    \\Annotator 2: For, Benevolent
    \\Consensus: Both, Hostile+Benevolent}
    \begin{quote}
    \small
    \textit{
    Sir A. STEEL-MAITLAND, 1923
    \\``... The Noble Lady the Member for the Sutton Division of Plymouth (Viscountess Astor) has stated the objects which the women police could effectually carry out, and may I add, from my own experience, that I can most heartily support what she has said? No one in their senses supposes that you can ask any woman to be a policewoman, whatever be her training or her temperament, or that you can ask them to do any sort of duties that the ordinary male constable does at the present time. They are useful, if they are well trained, for special kinds of work, and special kinds only, but speaking, not as a sentimentalist, but merely in order to get the ordinary proper police work carried out well, there are some kinds of police work which can better be effected by trained women than by any male constable that exists. The hon. Member for the Sutton Division has referred to some of them. There is the question of questioning small children when they are brought up for examination, escorting prisoners, watching suicides, and dealing with offences of soliciting in the streets...'' 
    }
    \end{quote}
    The speaker supports the representation of women in the police force but only in so far as dealing with children, street workers, etc. Thus, the consensus label reflects these ambivalent sentiments with stance as \textit{both} and sexism as \textit{Competitive and Complementary Gender Differentiation}.

    \item \textbf{Annotator 1: Against, Hostile
    \\Annotator 2: For, None
    \\Consensus: For, None}
    \begin{quote}
    \small
    \textit{
    Mr. F. E. SMITH, 1913
    \\``... What is the position in which the supporters of this movement find themselves to-day? They have no majority among the male voters. This is a point upon which it is very easy to dogmatise. It is common knowledge that no one dare fight an election on Female Suffrage, and no one will. If you get anyone to do it he is beaten by four thousand votes ...'' 
    }
    \end{quote}
    The speaker says the members would find it difficult find support for Female Suffrage in their respective constituencies. This can be interpreted as being for or against suffrage. Finally, the consensus label upon discussion gives the benefit of the doubt labelling stance as \textit{For} and sexism as \textit{None}.

    \item \textbf{Annotator 1: For, Benevolent
    \\Annotator 2: Irrelevant, None
    \\Consensus: For, Benevolent}
    \begin{quote}
    \small
    \textit{
    Mr. Bevin, 1941
    \\``... I have arranged that women doctors will serve on medical boards—as is the practice now when women are examined, but there has not been the amount of examination that will develop under this arrangement ... We shall give a woman the opportunity of going into one of the Auxiliary Services, certain duties in the Civil Defence, or certain specified jobs in industry. This gives an opportunity to women who, for various reasons, may not be able to go away, but who can volunteer for certain forms of Civil Defence in their own localities. Also it gives an opportunity to the woman who prefers industry ... We have decided, after all the representations, to exclude the married women from compulsory military service, though they will be subject to direction to employment. It makes a great deal of difference to the psychology of the men who are in the Forces, especially overseas, and we cannot ignore it. The childless married woman can be directed to work, but we are going to pay very great attention to the women with children and see that directions are not issued to direct them away from their homes ...'' 
    }
    \end{quote}
    The speaker advocates for the participation of women multiple public spheres but limited to certain roles, and excludes married women from the Civil Defence sector out of protectiveness. This could have been irrelevant to the topic as well as not-sexist but upon discussion, women in the different job roles were considered part of public representation and exclusion, despite intention, was considered sexist. The consensus label was then \textit{For} in terms of stance and \textit{Benevolent-Protective Paternalism \& Complementary Gender Differentiation} for sexism.
    
\end{itemize}

\subsection{Human-LLM (Claude Sonnet 4.6) Disagreement}

Claude Sonnet 4.6 reaches Cohen's $\kappa = 0.711$ against the human-consensus labels, comparable to the two annotators’ agreement with each other ($\kappa = 0.644$). The following examples illustrate some interesting cases:

\begin{itemize}
    \item \textbf{LLM: Both, Benevolent
    \\Human Consensus: Against, Hostile+Benevolent}
    \begin{quote}
    \small
    \textit{
    Mr. HUGH LAW, 1912
    \\``... We were not elected on the question of Woman Suffrage, we were elected to advance Home Rule, and we are bound to make that our first consideration ... I speak quite frankly—though I suppose in a declaration of this sort I may be under suspicion in present circumstances—and say that, like my hon. Friend the Member for East Clare (Mr. W. Redmond) I am a convinced Suffragist. I look upon this question not as one in which men are arrayed against women. I look upon it primarily as an Irish question, and I ask myself, What is it that Irish women themselves desire? From that point of view I confess I think there can be only one answer. I am bound to answer that, if I look to the desires of Irish women themselves in the matter, I cannot be constrained to believe there is among them any such demand for the exercise of the suffrage as ought to cause us to deflect our policy ... If I believed that such a proportion of Irish women wore in favour of this proposal, as I believe are in favour of Woman Franchise in England, I should find myself in a different position... When I say "effective demand"—the words have been used more than once—I do not mean a militant demand... when Irish women ask for the vote, I will not deny it ...'' 
    }
    \end{quote}
    The speaker emphasizes that he will not vote against suffrage but still argues there is no demand for and will therefore vote against it at this time. He also talks about the kind of demand he will respect, i.e., \textit{``not militant''}. While the humans feels the supportive statements are for appeasement rather than sincere, the LLM labels it as \textit{Both} for stance. For sexism, humans label it \textit{Hostile and Benevolent Paternalism}, the LLM labels it \textit{Benevolent Gender Differentiation}.

    \item \textbf{LLM: Irrelevant, None
    \\Human Consensus: Against, Hostile}
    \begin{quote}
    \small
    \textit{
    Mr. GOLDMAN, 1913
    \\``... The two subjects in the Bill, although correlated, are distinct and unconnected in themselves ... Registration and the franchise have not only been looked upon as fundamentally distinct subjects; they have been regarded as problems too large to be legitimately contained within the area of one single Bill ... I do not wish to criticise the Bill at this stage. My whole object in mentioning this point is merely to point out the vast scope of this Bill and to say that it seems to me unfair and irrational that vast changes involving such vast expenditure should be passed and carried in this House without ample opportunity for discussion ... They will either be compelled to vote against what they, in their conscience, believe to be imperatively necessary for the proper representation of the people or they will be compelled to accept Amendments which they sincerely believe to be fraught with great danger to the country ... In my Constituency, for instance, a supporter of mine was disfranchised because during a period of illness his wife had received a single loaf from a relieving officer, and, in another case, a lodger who was living with his widowed mother and wanted to change into a residential qualification, lost his vote and was disfranchised. It is not only that...'' 
    }
    \end{quote}
    The speaker wishes to separate the question of Suffrage from the current bill as he feels it is a distinct topic. It can be looked upon as a delaying tactic as he does feel strong about male disenfranchisement and thus, the humans label it \textit{Against} and \textit{Hostile Paternalism}. The LLM labels it as \textit{Irrelevant} for stance citing the rationale as focusing on procedure, precedent and legislative efficiency.

    \item \textbf{LLM: For, None
    \\Human Consensus: Against, Hostile}
    \begin{quote}
    \small
    \textit{
    Sir H. CROFT, 1933
    \\``I congratulate the Government on their latest champion, who with her usual fearlessness has entered the lists. I hope the Government may take seriously what the Noble Lady has said. I am inclined to think that if there was a general demand in India that all women should have the vote, that is a question which might suitably be referred back again to India, and in the meantime the Joint Committee might adjourn in order to see what the Indian people think about such a suggestion. The Noble Lady would be content, I believe, with one vote for every woman. '' 
    }
    \end{quote}
    The speaker pushes for Indian Women's Franchise to be referred back to colonial Indian legislators rather than decide in the current bill. He also adds a sarcastic remark suggesting Viscountess Astor would ask for plural votes and would not be satisfied with a regular vote. Hence, the humans label it \textit{Against} and \textit{Hostile Paternalism \& Gender Differentiation}. The LLM labels rationalizes it as positive engagement and support of the Viscountess and tags it as \textit{For} in stance without any sexism detected.
\end{itemize}

\section{Classification Reliability}
\label{sec:classification_reliability}

While the sexism classification is an inherently subjective and complex task, we agree that Cohen's $\kappa$ less than 0.4 is quite low. This reflects the difficulty of the task and is consistent with prior work on sexism annotation in contemporary text \citep{kirk-etal-2023-semeval, guest-etal-2021-expert}. Long-form and archaic rhetoric adds to this complexity. To address uncertainty, combined with the LLM's low recall (0.43--0.46), we re-ran the stance--sexism analysis using only the human consensus labels on the 300-speech subset since these labels do not depend on the LLM (Table~\ref{tab:sexism_by_stance_val}).

\begin{table}[h]
\centering
\footnotesize
\begin{tabular}{@{}lrrr@{}}
\toprule
& \textbf{For} & \textbf{Against} & \textbf{Both} \\
& \textit{(n=96)} & \textit{(n=30)} & \textit{(n=13)} \\
\midrule
No sexism                & 62.5\% & 23.3\% & 46.2\% \\
Hostile only             & 1.0\%  & 33.3\% & 7.7\%  \\
Benevolent only          & 34.4\% & 6.7\% & 23.1\% \\
Hostile and benevolent   & 2.1\%  & 36.7\% & 23.1\% \\
\bottomrule
\end{tabular}
\caption{Sexism categories by stance on 300 human-annotated speeches.}
\label{tab:sexism_by_stance_val}
\end{table}

On human labels, 76.7\% of Against-speeches contain sexism (23/30) versus 37.5\% of For-speeches (36/96). This gap is statistically significant (Fisher's exact test, p $<$ 0.001), and we find that the type of sexism also splits by stance as in the main results: 91.7\% of sexist For-speeches are benevolent-only (33/36), while 91.3\% of sexist Against-speeches involve hostility (21/23). We find, therefore, that the same pattern appears in labels produced independently of the LLM, which suggests the finding is not primarily an artifact of LLM under-detection.

\paragraph{\textbf{Implications of the analysis:}}
\begin{enumerate}
\item \textbf{Absolute rates are likely underestimates:} Claude's precision is high (0.77--0.86) but recall is low (0.43--0.46), so flagged speeches are generally reliable while a substantial fraction of sexist speeches is likely missed. This suggests the true rates are higher than the 21\% (For) and 54\% (Against) we report. The human-label rates are 37.5\% and 76.7\%: almost exactly what our precision/recall estimates would predict for the For class, and directionally consistent for Against. Benevolent sexism also appears harder to detect than hostile sexism (see Limitations). To the extent this biases our estimates, it would undercount benevolent sexism in For-speeches, meaning our finding that pro-suffrage sexism is predominantly benevolent is likely understated rather than inflated.

\item \textbf{Relative comparisons appear more stable:} Our main claims compare stances with types of sexism. Under-detection lowers both sides of these comparisons, and unless misses are strongly concentrated in one stance, it should not reverse them. Against-speeches prove to be 2.0x more sexist on human labels and 2.6x on corpus labels, showing similar contrasts.
\end{enumerate}

\section{Distribution of Irrelevant Speeches}
\label{sec:irrelevant_distribution}

Systematic differences in the type of speeches could influence the composition of the final corpus and, consequently, the downstream temporal analyses. Therefore, upon examining whether the irrelevant classifications are distributed uniformly across time periods and debate contexts, we found that the irrelevant rate is not uniform over time but varies predictably as shown in Table~\ref{tab:irrelevant_distribution}.

    \begin{table}[h]
    \centering
    \footnotesize
    \begin{tabular}{@{}lrr@{}}
    \toprule
    \textbf{Era} & \textbf{n} & \textbf{Irrelevant} \\
    \midrule
    pre-1870 & 252 & 84.5\% \\
    1870--1899 & 775 & 41.7\% \\
    1900--1918 & 1680 & 42.6\% \\
    1919--1928 & 676 & 44.7\% \\
    1929--1950 & 957 & 65.9\% \\
    post-1950 & 2191 & 64.1\% \\
    \bottomrule
    \end{tabular}
    \caption{Distribution of Irrelevant speeches across eras.}
    \label{tab:irrelevant_distribution}
    \end{table}

The irrelevant rate is highest where keyword false positives are most plausible --- before 1870, when ``franchise'' typically referred to male franchise reform (84.5\%), and after equal franchise was settled in 1928 (64--66\%) --- while retention in the core suffrage era (1870--1928) is stable at 42--45\%, suggesting differential filtering is unlikely to drive our temporal results. Consistent with this, high-precision Tier~1 keywords are discarded less often than Tier~2 (48\% vs.\ 60\%), and a manual audit found 30/30 random discards correctly classified.

\section{Speaker--Level Aggregation Analysis}
\label{sec:speaker_aggregation}

    \begin{table}[h]
    \centering
    \footnotesize
    \begin{tabular}{@{}L{0.2\textwidth}L{0.06\textwidth}L{0.06\textwidth}L{0.075\textwidth}@{}}
    \toprule
    \textbf{Result} & \textbf{Per-speech} & \textbf{Per-speaker} & \textbf{Speakers} \\
    \midrule
    Support (For) among female MPs' speeches & 92.8\% & 92.5\% & 168 \\
    Support (For) among male MPs' speeches  & 70.1\% & 70.3\% & 1,158 \\
    Any sexism in Against-speeches & 54.0\% & 48.9\% & 344 \\
    Any sexism in For-speeches & 21.3\% & 22.7\% & 1,104 \\
    Hostile sexism in Against-speeches & 43.5\% & 38.6\% & 344 \\
    Benevolent-only share in For-speeches  & 81.0\% & 81.4\% & 354 \\
    \bottomrule
    \end{tabular}
    \caption{Headline results counting each speech once (Per-speech) vs.\ counting each speaker once (Per-speaker: each speaker's own rate, averaged with equal weight per speaker). The final column gives the number of speakers behind each estimate.}
    \label{tab:speech_vs_speaker}
    \end{table}
    
The corpus does not appear to be dominated by prolific speakers since the 2,942 relevant speeches come from 1,429 unique speakers, the median speaker contributes 1 speech (maximum 32), and the 10 most prolific speakers together account for 7.7\% of speeches. The main findings hold when each speaker counts once, so that prolific speakers get no extra weight (Table~\ref{tab:speech_vs_speaker}): support for suffrage by gender is nearly unchanged, and opponents of suffrage remain roughly twice as likely as supporters to use sexist rhetoric (48.9\% vs.\ 22.7\%). The decline of hostile sexism across eras also holds.

\end{document}